\documentclass{bmvc2k}

\title{Dual-Pixel Raindrop Removal}
\usepackage{multirow}
\usepackage{booktabs}
\usepackage{bigstrut}

\addauthor{Yizhou Li}{yli@ok.sc.e.titech.ac.jp}{1}
\addauthor{Yusuke Monno}{ymonno@ok.sc.e.titech.ac.jp}{1}
\addauthor{Masatoshi Okutomi}{mxo@ctrl.titech.ac.jp}{1}

\addinstitution{
 Tokyo Insititute of Technology,\\
 Tokyo, Japan
}

\runninghead{Li, Monno, Okutomi}{Dual-Pixel Raindrop Removal}

\begin{document}

\maketitle

\begin{abstract}
Removing raindrops in images has been addressed as a significant task for various computer vision applications. In this paper, we propose the first method using a Dual-Pixel~(DP) sensor to better address the raindrop removal. Our key observation is that raindrops attached to a glass window yield noticeable disparities in DP's left-half and right-half images, while almost no disparity exists for in-focus backgrounds. Therefore, DP disparities can be utilized for robust raindrop detection. The DP disparities also brings the advantage that the occluded background regions by raindrops are shifted between the left-half and the right-half images. Therefore, fusing the information from the left-half and the right-half images can lead to more accurate background texture recovery.
Based on the above motivation, we propose a DP Raindrop Removal Network~(DPRRN) consisting of DP raindrop detection and DP fused raindrop removal. To efficiently generate a large amount of training data, we also propose a novel pipeline to add synthetic raindrops to real-world background DP images.
Experimental results on synthetic and real-world datasets demonstrate that our DPRRN outperforms existing state-of-the-art methods, especially showing better robustness to real-world situations.
Our source code and datasets are available at http://www.ok.sc.e.titech.ac.jp/res/SIR/.

\end{abstract}

\section{Introduction}
\label{sec:intro}

Raindrops are typically attached to a glass window or a windshield and refract the light from the scene similar to a fish-eye lens. Therefore, raindrops in images eliminate the original background textures on raindrop-covered regions, which greatly reduces the image visibility and may disturb many computer vision tasks, e.g., object detection~\cite{objectdetection}, video surveillance~\cite{videosur}, and autonomous driving~\cite{deepdriving}. 
To avoid these disadvantages, removing raindrops in images has been treated as one of the important low-level vision tasks.


Recently, many deep-learning-based methods have been proposed to address single image deraining, including rain streak removal~\cite{DDN, jorder, rescan, prenet, RCD-Net, spanet, ecnet} and raindrop removal~\cite{attgan, raindropSDA2019, raindropselective, raindroplaplacian, raindropuncertainty, raindropsyn2019}. Regarding the raindrop removal, representative methods include raindrop-mask-guided methods~\cite{attgan, raindropuncertainty, raindropselective, raindropsyn2019}, an edge-guided method~\cite{raindropSDA2019}, and a Laplacian-pyramid-based method~\cite{raindroplaplacian}, as well as general image restoration frameworks~\cite{mprnet, restormer, dgunet} validated to be effective on the raindrop removal. 
Despite the fact that these state-of-the-art single image raindrop removal methods achieve good performance on synthetic datasets, they often show degraded and limited performance on real-world data, as experimentally pointed out in the survey paper of~\cite{deraininggeneralizationsurvey}, because of the domain gap between synthetic training data and real-world data. 
One inherent challenge of the single image raindrop removal is that raindrop detection and removal highly rely on raindrop appearance~(e.g., raindrop textures and  shapes) observed in a single image. Therefore, they inevitally fail when there are large appearance gaps of raindrops between the synthetic training data and the real-world data at an application phase. 



\begin{figure*}[t!]
\begin{center}
\includegraphics[width=1.0\linewidth]{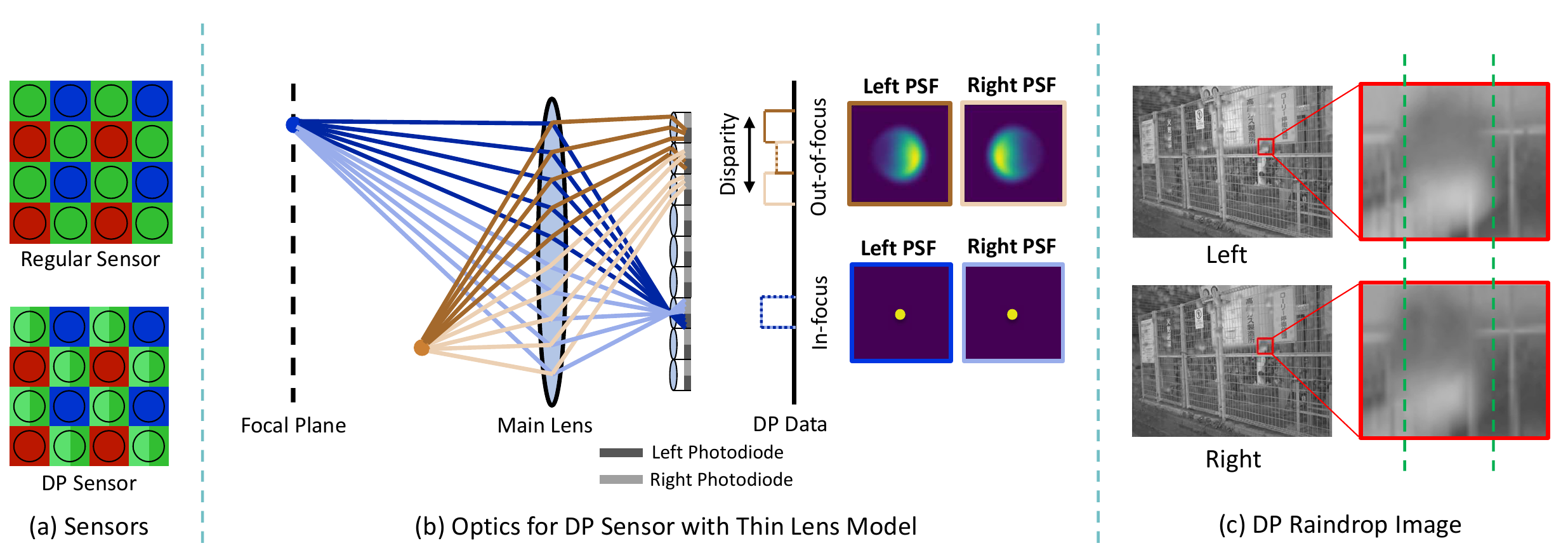}
\end{center}
    \vspace{-4mm}
    \caption{(a)~In a DP sensor (Google Pixel 4), each green pixel is split into two halves. (b)~The light rays from an in-focus point reach the same pixel, while the light rays from an out-of-focus point reach different pixels in the left-half and the right-half images, producing a DP disparity. The corresponding point spread functions~(PSFs), which are used to model the DP disparity, are shown in the right. (c)~Raindrops are usually out-of-focus and exhibit DP disparities, while in-focus backgrounds show almost no DP disparity. This motivates us to exploit the DP disparity for raindrop detection and removal.}
    \vspace{-4mm}
\label{fig:dp_optics}
\end{figure*}

In this paper, we focus on a Dual-Pixel~(DP) sensor to better address the raindrop removal task. A DP sensor has been adopted in some consumer digital cameras (e.g., Canon 5D Mark IV and Canon EOS R5) and smartphones (e.g., Google Pixel series and Samsung Galaxy series), as it can be implemented without significantly increasing the cost.
As shown in Fig.~\ref{fig:dp_optics}(a), a DP sensor divides each sensing pixel into two halves with left and right photodiodes, by which two individual images called left-half and right-half images can be captured. The summation of these two images, which we call a combined image, corresponds to the image captured by a regular sensor. As shown in Fig.~\ref{fig:dp_optics}(b), for an in-focus scene point within the Depth-of-Field~(DoF), the light rays reach the same pixel. In contrast, for an out-of-focus scene point outside the DoF, the light rays reach different pixels, leading to an intensity shift on the horizontal axis between the left-half and the right-half images, which is called a DP disparity. The studies~\cite{dpdispiccp, dpdeblur2020, dpdeblursyn2021, dpdebluropt2021} show that the DP disparity can be modeled as different Point Spread Functions~(PSFs) of the left-half and the right-half images as shown in Fig.~\ref{fig:dp_optics}(b), where the left-half and the right-half images can be generated from an all-in-focus image by convolution using different disk blur kernels corresponding to their varied PSFs.
Recent studies have also demonstrated that the DP disparity is useful for a wide range of applications, such as defocus deblurring~\cite{dpdeblur2020, dpdeblursyn2021, dpdebluropt2021}, autofocusing~\cite{autofocus1,autofocus2, autofocus3}, depth estimation~\cite{dpdisp2019, dpdispiccp, du2net}, joint deblurring and depth estimation~\cite{dpjointdepthdeblur}, and reflection removal~\cite{dpreflection}. However, to the best of our knowledge, there is no existing study that adopts a DP sensor for the raindrop removal. Although some of recent rain streak or raindrop removal studies exploit multi-view observations in a stereo or light-field setup \cite{lightfieldinpaint, lightfieldrainstreak, stereoderaining}, the DP sensor has the advantage that it can capture two perfectly aligned and synchronized images in one shot.



A key observation to motivate us to adopt a DP sensor is that, in practical applications requiring raindrop removal such as autonomous driving, the camera is usually background-focused, while the raindrops which are closer to the camera are out-of-focus. Therefore, as shown in Fig.~\ref{fig:dp_optics}(d), the raindrop regions exhibit noticeable DP disparities while other background regions show almost no DP disparity. This suggests that we can utilize DP disparities, instead of solely depending on the raindrop appearance, to detect raindrop locations. Since the DP disparities for raindrops bring slightly-shifted raindrop positions in the left-half and the right-half images, the non-raindrop-covered visible background textures in those images are also shifted. This suggests that we can exploit this redundancy to better restore the background details by fusing the information from the leaf-half and the right-half images.

With the above motivation, we propose DP Raindrop Removal Network~(DPRRN), which consists of two main parts: DP raindrop detection and DP fused raindrop removal. Firstly, the DP raindrop detection part robustly predicts the raindrop masks on the left-half and the right-half images, according to the DP disparities of raindrops. Then, the DP fused raindrop removal part removes the raindrops with the predicted raindrop masks as guide, where we intermediately remove the raindrops from the left-half and the right-half images, respectively, and then fuse and refine those results to obtain the final raindrop removal result as a combined image, i.e., an image captured by a regular sensor. The networks for both parts are trained in an end-to-end manner.

Since existing raindrop removal datasets are not applicable to a DP sensor, we built the first DP raindrop removal dataset. To efficiently generate a large amount of training data, we present a novel pipeline to add synthetic raindrops to real-world background DP images. We captured 613 real-world scenes and generated 2,452 pairs of training data with synthetic raindrops. In addition, to test the generalization performance in real-world situations, we also collected a real-world dataset containing 82 DP image pairs with and without real raindrops, which contains raindrop patterns and background textures that are unseen in the training data. Main contributions of this work are summarized as follows:
\begin{itemize}
\vspace{-2mm}
    \item We propose the first DP raindrop removal network and dataset. Our network effectively utilizes DP disparities on raindrops for raindrop detection and removal. 
\vspace{-2mm}
    \item We experimentally demonstrate that our DP raindrop removal network achieves state-of-the-art performance and shows better robustness to real-world situations. 
\end{itemize}

\section{DP Raindrop Removal Datasets}
\label{sec:dataset}

\begin{figure*}[t!]
\begin{center}
\includegraphics[width=1.0\linewidth]{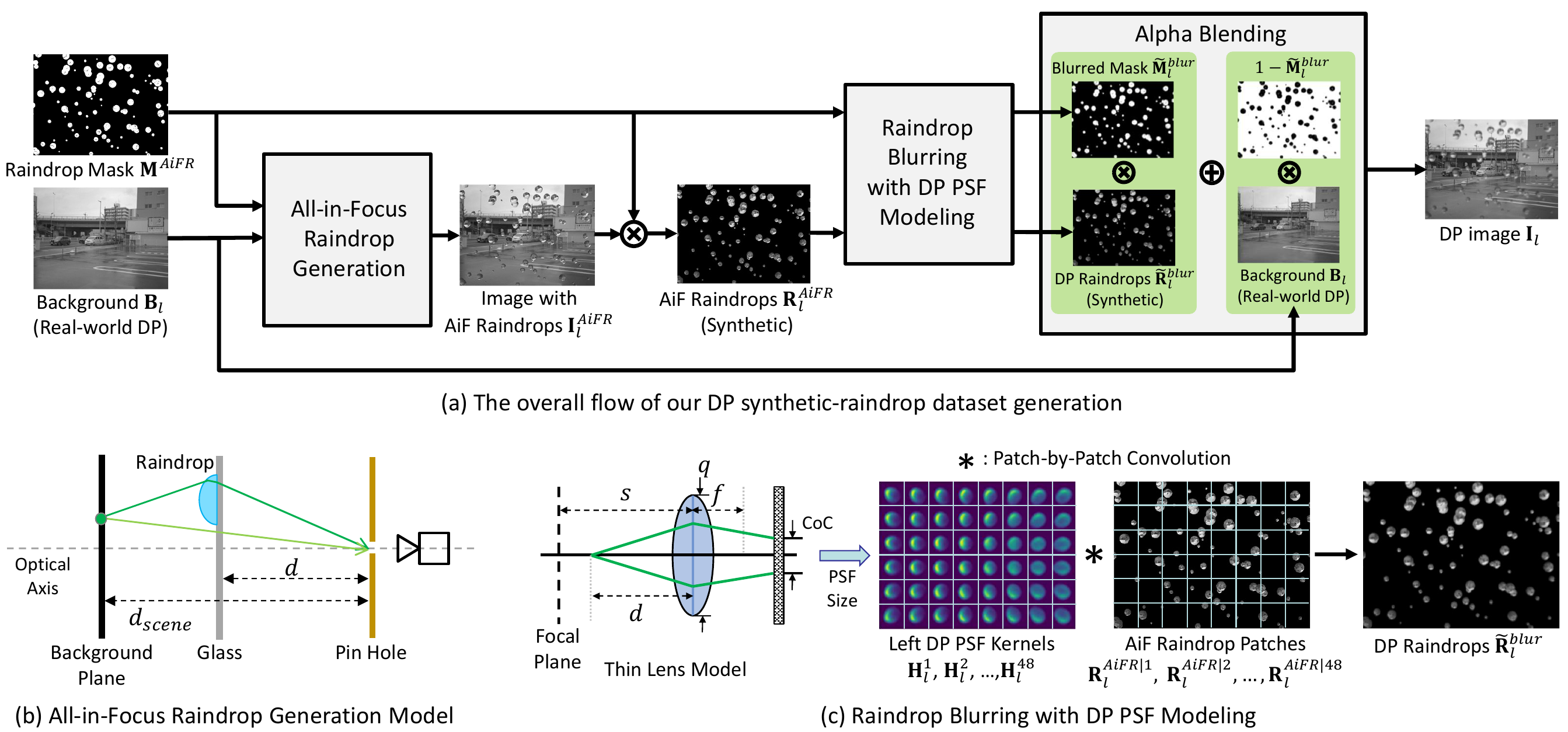}
\end{center}
    \vspace{-4mm}
    \caption{Proposed DP synthetic-raindrop dataset generation: (a) The overall flow for the left-harf image, which consists of raindrop generation, raindrop blurring, and alpha blending. (b) A pin-hole model to generate all-in-focus raindrops. (c) The raindrop blurring process with DP PSF modeling. The similar process is conducted for the right-half image.}
\label{fig:dp_render}
\end{figure*}

\subsection{Synthetic-Raindrop Dataset}
\label{ssec:synthetic}

It is laborious to collect a large number of real-world aligned image pairs with and without real raindrops. It is also hard to accurately simulate DP sensors' inherent characteristics, such as uneven vignetting for the left-half and the right-half images. Therefore, to construct a training dataset for our network, we take a hybrid approach, which only synthesizes raindrops and adds them to real background DP images, which can be easily captured using a real DP camera as usual. In all our experiments, we used Google Pixel~4. Since Google Pixel~4 adopts the DP sensor architecture only to the green pixels, we used green-channel images for our dataset construction, though our dataset construction and network architecture can be easily extended to the RGB color domain.


Figure~\ref{fig:dp_render}(a) shows the proposed data generation pipeline, where the processes for the left-half image are presented. The processes for the right-half image can similarly be performed. The pipeline consists of three main parts: all-in-focus raindrop generation, raindrop blurring, and alpha blending of synthetic raindrops and a real background. 

The first part generates all-in-focus raindrops based on the raindrop refraction model of~\cite{raindropsyn2019}. For simplicity, as shown in Fig.~\ref{fig:dp_render}(b), we assume that the whole background has a constant depth $d_{scene}$ and the raindrops (glass) has another smaller constant depth $d$. We add all-in-focus raindrops to the real-world background DP image ${\bf B}_l$ with ray tracing to generate the image with all-in-focus raindrops ${\bf I}_l^{AiFR}$. The locations of the raindrops are determined by a randomly assigned binary raindrop mask ${\bf M}^{AiFR}$, where ${\bf M}^{AiFR}(x)=1$ means that the pixel $x$ is a part of a raindrop region and ${\bf M}^{AiFR}(x)=0$ means that the pixel is in background regions. We set the background depth~$d_{scene}=10m$ and randomly change the raindrop depth~$d$ between $15cm$ and $25cm$. 


The second part simulates DP disparities for the raindrops. For this purpose, we apply the spatially-varying DP PSF modeling in~\cite{dpdebluropt2021}, where in total $6\times 8$ PSF kernel shapes, denoted as ${\bf H}_l^i$ for the left-half image, are calibrated for each sub-patch $i$, as shown in Fig.~\ref{fig:dp_render}(c). To determine the scale of PSF kernels, we calculated the Circle of Confusion~(CoC) radius $r=\frac{q}{2}\times \frac{s'}{s} \times \frac{d-s}{d}$, which corresponds to the PSF radius, where $q$ is the aperture diameter, $s'$ is the distance between the lens and the sensor, $s$ is the focus distance, and $d$ is the assumed raindrop depth. Specifically, using a thin lens model, $q$ is calculated as $q=\frac{f}{F}$ according to Google Pixel 4's focal length $f$ and F-stop $F$, and $s'$ is calculated as $s'=\frac{fs}{s-f}$, where the focus distance is set as $s=d_{scene}$ because the camera is assumed to be background-focused. According to the calculated CoC radius, we re-scale the PSF kernels to derive ${\bf H}_{l}^i$ in correct PSF size for the real camera.

To blur the all-in-focus raindrops with the above PSF kernels, the raindrops ${\bf R}_l^{AiFR}$ are extracted from ${\bf I}_l^{AiFR}$ as ${\bf R}_l^{AiFR}= {\bf M}^{AiFR} \otimes {\bf I}^{AiFR}_l$, where $\otimes$ denotes pixel-wise multiplication.
Then, the raindrops ${\bf R}_l^{AiFR}$ are divided into $6\times8$ patches and convolved with the PSF kernels in a patch-by-patch manner as $\Tilde{\bf R}_{l}^{blur|i}={\bf H}_{l}^i * {\bf R}_l^{AiFR|i}$, where $*$ denotes the convolution operation. The sub-patches are then stitched to form the blurred raindrop image $\Tilde{\bf R}_l^{blur}$.
Similarly, the binary raindrop mask ${\bf M}^{AiFR}$ is also blurred to generate the blurred raindrop mask $\Tilde{\bf M}_l^{blur}$, which is used as the weight for the alpha-blending in the next step.

The final part performs the alpha blending of the synthetic raindrops and the real-world background as ${\bf I}_{l}=\Tilde{\bf M}_{l}^{blur} \otimes \Tilde{\bf R}_{l}^{blur} + ({\bf 1}-\Tilde{\bf M}_{l}^{blur}) \otimes {\bf B}_l$ to generate the left-half image ${\bf I}_l$ with smooth and natural transitions from the raindrops to the background.
The similar process is conducted to generate the right-half image ${\bf I}_{r}$, where the corresponding right-half background image ${\bf B}_r$ and DP PSF kernels ${\bf H}_r^i$ are used for the same all-in-focus raindrop mask ${\bf M}^{AiFR}$. For the combined images corresponding to the images captured by a regular sensor, we generate them by averaging the left and the right images as ${\bf I}_c=\frac{{\bf I}_l+{\bf I}_r}{2}$ and ${\bf B}_c=\frac{{\bf B}_l+{\bf B}_r}{2}$.

\begin{figure*}[t!]
\begin{center}
\includegraphics[width=1.0\linewidth]{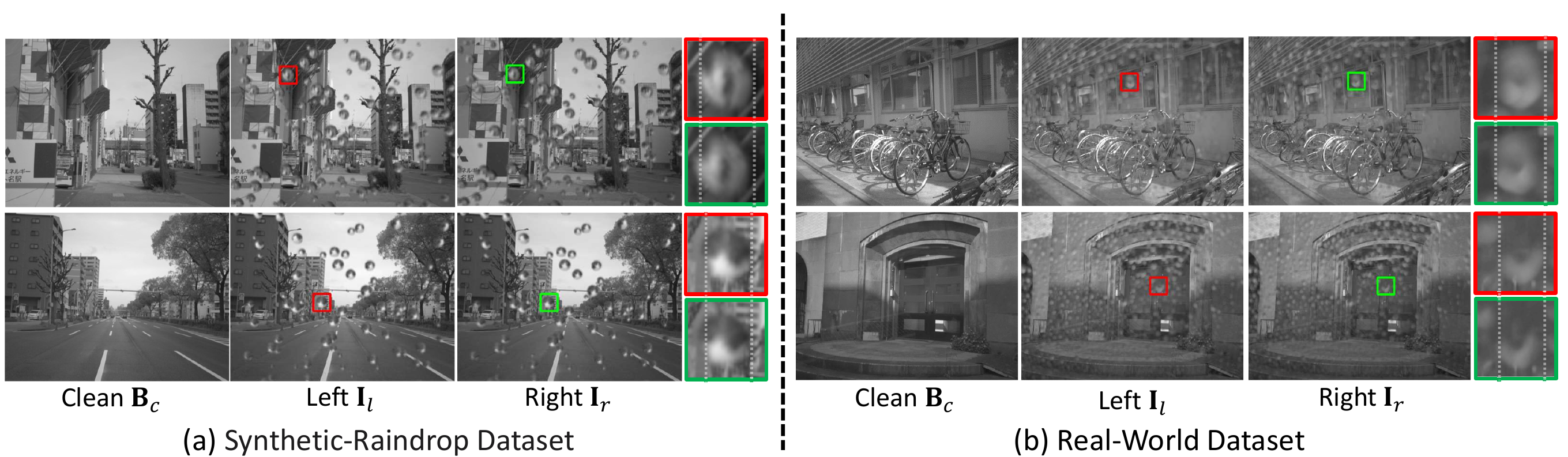}
\end{center}
    \vspace{-4mm}
    \caption{Constructed DP raindrop removal datasets: (a) Synthetic-raindrop dataset, which consists of a real background and synthetic raindrops. (b) Real-world dataset, which consists of both a real background and real raindrops. As depicted in the red and the green boxes, we can successfully simulate the DP disparities on raindrops in our synthetic-raindrop dataset.}
    \vspace{-4mm}
\label{fig:dp_data_sample}
\end{figure*}

For each real-world background DP image pair, we generate 4 different synthetic-raindrop image pairs by randomly changing the raindrop depth and mask. Each data contains the generated images of ${\bf I}_l$, ${\bf I}_r$, ${\bf I}_c$ and the ground-truth background images of ${\bf B}_l$, ${\bf B}_r$, ${\bf B}_c$. We captured 613 scenes and generated 2,452 image pairs in total, among which 1,960 pairs are used for training and the rest 492 pairs are used for testing. Two samples are shown in Fig.~\ref{fig:dp_data_sample}(a), where noticeable DP disparities exist in the raindrop regions.


\subsection{Real-World Dataset}
\label{ssec:realworld}
To evaluate the generalization performance, we also collected 82 real-world DP image pairs with and without real raindrops. To minimize the impact of background refraction changes, we only used one glass panel and fixed it at a position of $15$-$25cm$ from the camera. Then, we first captured a DP image pair with the clean panel as the ground-truth images. Then, we sprayed water onto the panel to generate real raindrops and captured another DP image pair as the input images. To reduce the impact of motion, we used a solid tripod to fix the smartphone and the panel, and remotely controlled the smartphone to shoot images. As shown in Fig.~\ref{fig:dp_data_sample}(b), we carefully collected 82 pairs of high-quality and well-aligned DP images with and without real raindrops, containing varied and complex raindrop patterns.



\section{Proposed DP Raindrop Removal Network}
\label{sec:method}

\begin{figure*}[t!]
\begin{center}
\includegraphics[width=1.0\linewidth]{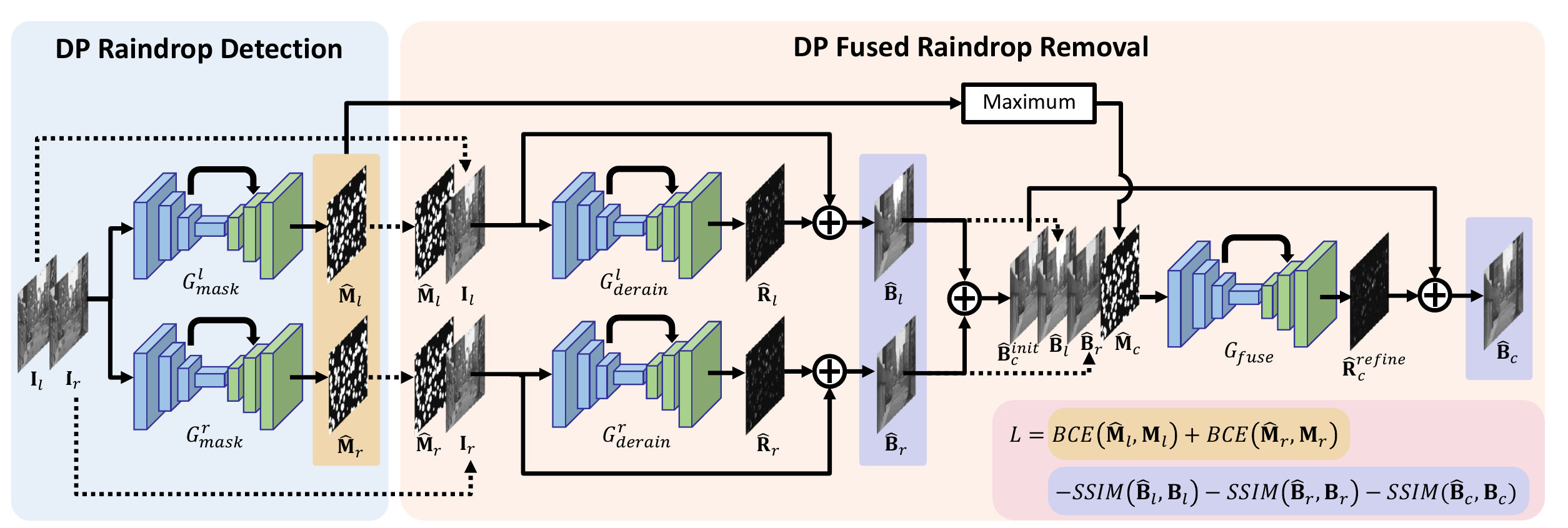}
\end{center}
    \vspace{-5mm}
    \caption{The overview of our proposed DP Raindrop Removal Network~(DPRRN).}
\label{fig:dp_network}
\end{figure*}

Figure~\ref{fig:dp_network} shows the overview of our proposed DP Raindrop Removal Network~(DPRRN). We input the pair of the left-half and the right-half images $\{{\bf I}_l, {\bf I}_r\}$ and estimate the clean background image $\hat{\bf B}_c$ as close as the ground truth ${\bf B}_c$, where the final output $\hat{\bf B}_c$ is a combined image as captured by a regular sensor. Our network consists of two parts: (i) DP raindrop detection, where we conduct raindrop detection on ${\bf I}_l$ and ${\bf I}_r$, respectively, and (ii) DP fused raindrop removal, where we remove raindrops on ${\bf I}_l$ and ${\bf I}_r$, respectively, whose results are subsequently fused and refined for the final raindrop removal result. Each network is constructed using UNet~\cite{unet} with residual blocks~\cite{he2016deep}, for which we refer to the supplementary material. All the networks are trained in an end-to-end manner, as detailed below.

\subsection{DP Raindrop Detection}

In DP raindrop detection, the input images ${\bf I}_l$ and ${\bf I}_r$ are channel-wise concatenated and sent to two different networks $G_{mask}^l$ and $G_{mask}^r$ to predict two raindrop masks $\hat{\bf M}_l$ and $\hat{\bf M}_r$ depicting the raindrop locations of ${\bf I}_l$ and ${\bf I}_r$, respectively. For the left-half mask $\hat{\bf M}_l$, the concatenated ${\bf I}_l$ and ${\bf I}_r$ are sent to $G_{mask}^l$ to predict a pixel-wise soft raindrop mask $\hat{\bf M}_l$ through Sigmoid activation function at the last layer. In this process, $G_{mask}^l$ performs raindrop localization guided with the DP disparities on the raindrops. Similarly, for the right-half mask $\hat{\bf M}_r$, another network $G_{mask}^r$ is utilized to predict a pixel-wise soft raindrop mask $\hat{\bf M}_r$. Here, $G_{mask}^l$ and $G_{mask}^r$ are two individual networks that do not share the parameters because the disparity of ${\bf I}_r$ to ${\bf I}_l$ and the disparity of ${\bf I}_l$ to ${\bf I}_r$ show inversed shift directions and we have found using two individual networks for them achieves more robust performance.

As for the loss function on the raindrop detection, we use the Binary Cross Entropy (BCE) \cite{bceloss} losses as
\begin{equation}
    L_{Mask}=BCE(\hat{\bf M}_l, {\bf M}_l) + BCE(\hat{\bf M}_r, {\bf M}_r),
\end{equation}
where ${\bf M}_l$ and ${\bf M}_r$ are the ground-truth binary raindrop masks, where the value one means that the corresponding pixel is a part of the raindrop regions of ${\bf I}_l$ and ${\bf I}_r$, respectively.

\subsection{DP Fused Raindrop Removal}
In DP fused raindrop removal, we first conduct raindrop removal of DP images ${\bf I}_l$ and ${\bf I}_r$ separately, with the guide of the estimated masks $\hat{\bf M}_l$ and $\hat{\bf M}_r$ as raindrop location clues. For the rain removal network $G_{derain}^l$, we input the left-half image ${\bf I}_l$ concatenated with the mask $\hat{\bf M}_l$ to derive the raindrop residuals $\hat{\bf R}_l$, from which the left-half background image is predicted as $\hat{\bf B}_l={\bf I}_l-\hat{\bf R}_l$. The raindrop removal of the right-half image ${\bf I}_r$ is performed in the same manner using another network $G_{derain}^r$ to predict the right-half background image $\hat{\bf B}_r$. Here, $G_{derain}^l$ and $G_{derain}^r$ are also two individual networks without sharing the parameters to address the different blur kernels applied in ${\bf I}_l$ and ${\bf I}_r$.


In the above processes, the background DP images $\hat{\bf B}_l$ and $\hat{\bf B}_r$ are predicted through $G_{derain}^l$ and $G_{derain}^r$ basically in a single image input manner. However, as the raindrop locations and the visible background information in ${\bf I}_l$ and ${\bf I}_r$ are slightly different according to the DP disparities, the restored background details in $\hat{\bf B}_l$ and $\hat{\bf B}_r$ are varied as well. This suggests the potential to derive a more accurate restoration result by fusing and refining the results of $\hat{\bf B}_l$ and $\hat{\bf B}_r$.

We then apply another network $G_{fuse}$ for fusing the raindrop removal results $\hat{\bf B}_l$ and $\hat{\bf B}_r$. Because our target is to predict a clean background image as captured by a regular sensor, we calculate the corresponding mask as $\hat{\bf M}_c=max(\hat{\bf M}_l, \hat{\bf M}_r)$ to derive a pixel-wise soft raindrop mask to the combined image ${\bf I}_c$, where $max$ denotes the pixel-wise maximum operation. We also calculate the initial combined background image as $\hat{\bf B}_c^{init}=\frac{\hat{\bf B}_l+\hat{\bf B}_r}{2}$. 
Then, the channel-wise concatenation of ${\hat{\bf M}_c}$, $\hat{\bf B}_l$, $\hat{\bf B}_r$, $\hat{\bf B}_c^{init}$ is sent to $G_{fuse}$ to derive the residuals $\hat{\bf R}_c^{refine}$ in terms of $\hat{\bf B}_c^{init}$, from which the final output background image is derived as $\hat{\bf B}_c=\hat{\bf B}_c^{init}-\hat{\bf R}_c^{refine}$. 

The DP fused raindrop removal is optimized using negative SSIM losses~\cite{prenet, BRN} to maximize SSIMs between the raindrop removal results and the ground-truth clean background images as
\begin{equation}
    L_{derain} = -SSIM(\hat{\bf B}_l, {\bf B}_l) - SSIM(\hat{\bf B}_r, {\bf B}_r) - SSIM(\hat{\bf B}_c, {\bf B}_c),
\end{equation}
which is the simple summation of three losses for each raindrop removal output.

Combined with the DP raindrop detection and the DP fused raindrop removal, the whole DPRRN is trained using the simple summation of the mask detection loss and the raindrop removal loss as 
\begin{equation}
    L = L_{mask} + L_{derain}. 
\end{equation}



\section{Experimental Results}
\label{sec:experiments}

\subsection{Implementation Details}
Our DPRRN is implemented using Pytorch~\cite{paszke2019pytorch} and trained on a single NVIDIA RTX3090 GPU. During the training, we use randomly cropped $480\times120$ patches with the batch size set to 12. To ensure that the shift directions of raindrops in the DP image pair do not change, we do not apply random flipping for data augmentation. ReLU~\cite{alexnet} is set as the activation function. RAdam~\cite{radam} is used as the optimizer to train the network with an initial learning rate of $1e^{-3}$. We adopt two-stage training, where we first train DP raindrop detection with $L_{mask}$ only for 100 epochs without changing the learning rate and then use all the losses $L=L_{mask}+L_{derain}$ to train the whole DPRRN in end-to-end for another 400 epochs, during which the learning rate is decayed by multiplying 0.2 at 120, 240, and 360 epochs.

\subsection{Comparison with State-of-the-Art Methods}
\textbf{Compared Methods}: We compared AttGAN~\cite{attgan} and CCN~\cite{ccn}, which are the methods designed for single image raindrop removal. Since AttGAN has not released the official training code, we referred to the past practice and used a third-party repository~\cite{attgangithub} to train AttGAN on our dataset. CCN is proposed to remove rain streaks and raindrops in one go. Since our current target is raindrop removal only, we took the raindrop removal part of CCN for training and comparison.
We also compared general image restoration methods, MPRNet~\cite{mprnet}, DGUNet~\cite{dgunet}, and Restormer~\cite{restormer}, which are very recent state-of-the-art methods. Although these methods are not specifically designed for raindrop removal, they all contain comparative experiments on raindrop removal and demonstrate their effectiveness. For all the above single image methods, the input is a regular single image, which is calculated as ${\bf I}_c=\frac{{\bf I}_l+{\bf I}_r}{2}$, and the output target is a clean background image in a regular sensor domain, which is supervised by the ground-truth background image calculated as ${\bf B}_c=\frac{{\bf B}_l+{\bf B}_r}{2}$. We used their default parameter settings to train the networks.

As a DP-based method, a recent DP-based defocus deblurring method RDPD+~\cite{dpdeblursyn2021} was compared to validate the effectiveness of our network design utilizing DP images. The input to both RDPD+~\cite{dpdeblursyn2021} and proposed DPRRN is a DP image pair ${\bf I}_l$ and ${\bf I}_r$, and the final output is also supervised by a clean background image ${\bf B}_c$ in a regular sensor domain. Since RDPD+ does not have specific design for defocusing deblurring, it can be directly trained on our DP synthetic-raindrop dataset. As an ablation study, we also compared our DPRRN without the DP raindrop detection part, which is denoted as DPRRN$^{\rm -RD}$, where the only DP fused raindrop removal part was trained without the guide of the raindrop masks.


\begin{table}
    \centering
    \caption{Quantitative comparison on our datasets. (Red: rank 1st; Blue: rank 2nd)}
    \aboverulesep=0ex
    \belowrulesep=0ex
    \label{tb:psnr}
    \renewcommand{\arraystretch}{1.0}
    \renewcommand\tabcolsep{5pt}
    \vspace{2.0mm}
    \begin{tabular}{c|l|cc|cc}
        \toprule
        \multirow{2}{*}{Input Image Types}             & \multicolumn{1}{c|}{\multirow{2}{*}{Methods}} & \multicolumn{2}{c|}{Synthetic} & \multicolumn{2}{c}{Real-World} \\
        \cmidrule{3-6}
                              &  & PSNR & SSIM & PSNR & SSIM \\
        \midrule
        \multirow{5}{*}{Regular} & AttGAN~(CVPR2018)~\cite{attgan} & 34.55 & 0.9614 & 29.59 & 0.9127        \\
                           & CCN~(CVPR2021)~\cite{ccn} & 38.88 & 0.9790 & 27.92 & 0.9001        \\
                           & MPRNet~(CVPR2021)~\cite{mprnet} & 41.02 & 0.9828 & 30.83 & 0.9247        \\
                           & Restormer~(CVPR2022)~\cite{restormer} & \textbf{\textcolor{blue}{41.94}} & \textbf{\textcolor{blue}{0.9845}} & 30.88 & 0.9248 \\
                           & DGUNet~(CVPR2022)~\cite{dgunet} & 41.32 & 0.9835 & 31.08 & 0.9259 \\
        \midrule
        \multirow{3}{*}{Dual Pixel} & RDPD+~(ICCV2021)~\cite{dpdeblursyn2021} & 40.51 & 0.9809 & 31.18 & 0.9205        \\
                           & \textbf{DPRRN$^{\bf-RD}$} (Ours)& 40.93 & 0.9824 & \textbf{\textcolor{blue}{31.85}} & \textbf{\textcolor{blue}{0.9357}}        \\
                           & \textbf{DPRRN} (Ours) & \textbf{\textcolor{red}{42.70}} & \textbf{\textcolor{red}{0.9867}} & \textbf{\textcolor{red}{32.47}} & \textbf{\textcolor{red}{0.9396}} \\
        \bottomrule
    \end{tabular}
\end{table}

\begin{figure*}[t!]
\begin{center}
\includegraphics[width=1.0\linewidth]{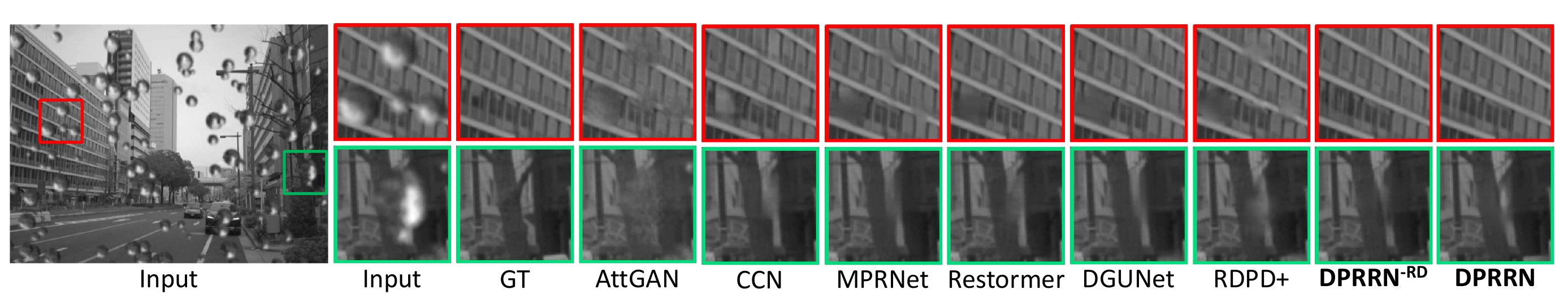}
\end{center}
    \vspace{-6mm}
    \caption{Qualitative comparison on the synthetic-raindrop dataset.}
    \vspace{-3mm}
\label{fig:syn_compare}
\end{figure*}

\vspace{1mm}
\noindent
\textbf{Results on Synthetic-Raindrop Dataset}: We here show the results, where both the training and the testing were performed on the synthetic-raindrop dataset as detailed in Sec.~\ref{ssec:synthetic}.
The second-right part of Table~\ref{tb:psnr} shows the quantitative results with PSNR and SSIM. The results show that our DPRRN achieves the best performance on the synthetic data. Compared with the past single image methods, our DPRRN achieves PSNR and SSIM improvements of 0.76dB and 0.0022, respectively. Compared with the DP-based method of RDPD+, our DPRRN also achieves PSNR and SSIM improvements with the large margins of 2.19dB and 0.0043, respectively. In comparison of DPRRN$^{\rm -RD}$ and DPRRN, we can confirm that the DP raindrop detection part significantly contributes to the performance improvement. These results validate that our DPRRN architecture can effectively exploit DP information to better remove raindrops. 
Figure~\ref{fig:syn_compare} shows the qualitative results on the synthetic dataset. For the raindrops in the red box, our DPRRN can restore the thin details of window frames, while the other methods exhibit over-smoothing results. For the raindrops in the green box, our DPRRN can recover sharper details around the branches compared with the other methods.


\begin{figure*}[t!]
\begin{center}
\includegraphics[width=1.0\linewidth]{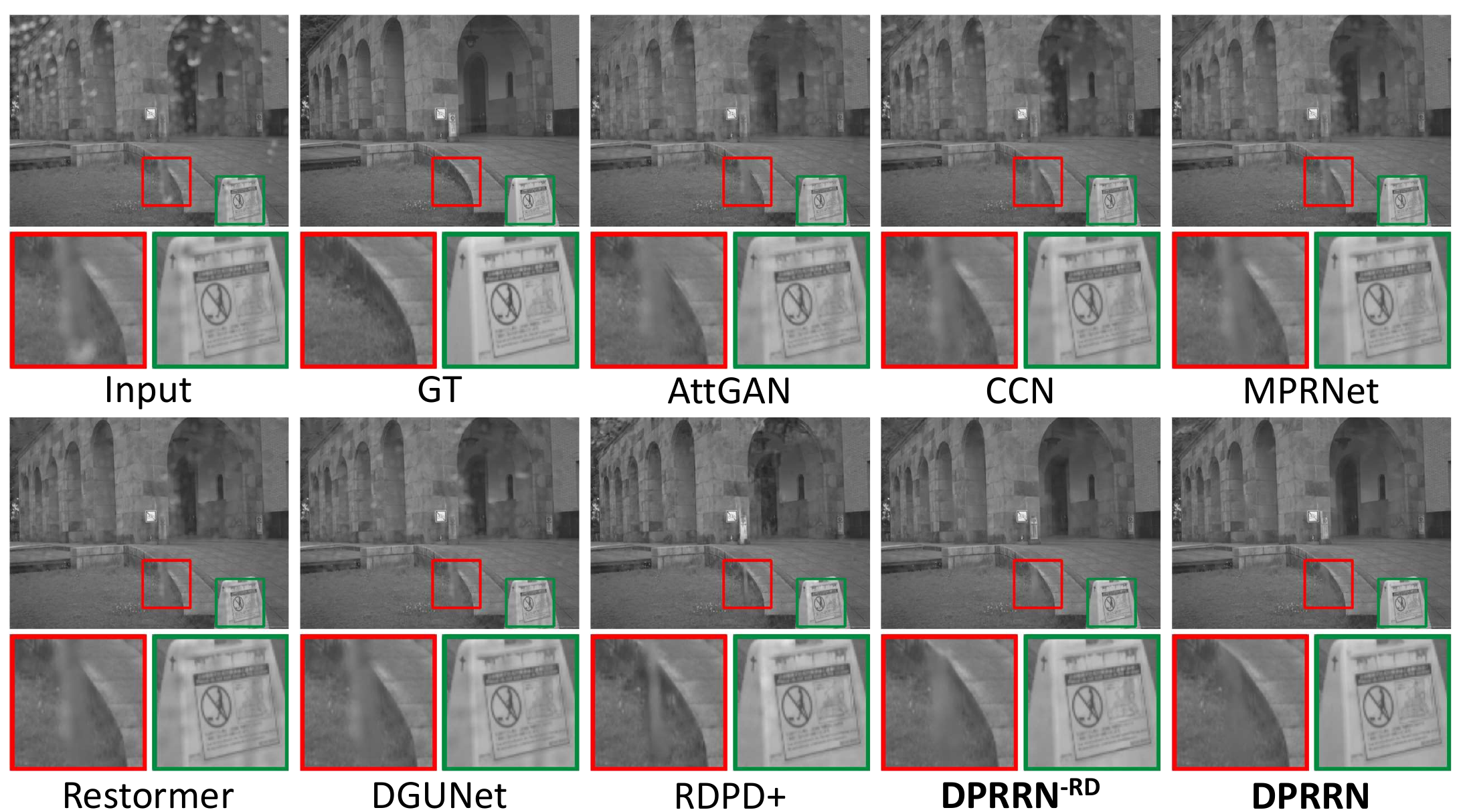}
\end{center}
    \vspace{-3mm}
    \caption{Qualitative comparison on the real-world dataset.}
    \vspace{-3mm}
\label{fig:real_compare}
\end{figure*}

\vspace{1mm}
\noindent
\textbf{Results on Real-World Dataset}: We next show the results, where the networks were trained using the synthetic-raindrop dataset and then tested on the real-world dataset as detailed in Sec.~\ref{ssec:realworld}. The rightmost part of Table~\ref{tb:psnr} shows PSNR and SSIM results, demonstrating that our DPRRN exhibits large PSNR and SSIM margins of 1.29dB and 0.0137 compared with the best-performed existing methods, respectively. These margins on the real-world data are larger than those on the synthetic data (i.e., 0.76dB and 0.0022), demonstrating higher generalization ability of our DPRRN. Although MPRNet, Restormer, and DGUnet show better results than DPRRN$^{\rm -RD}$ on the synthetic data, they show worse results than DPRRN$^{\rm -RD}$ on the real-world dataset. This indicates that, even without the raindrop detection part, our DPRRN shows better robustness to real-world situations.

Figure~\ref{fig:real_compare} shows the qualitative results on the real-world dataset. The existing methods have a large number of remaining raindrops, while our DPRRN is able to remove those raindrops and restore clean and sharp background details. Especially for the red and the green box regions, our DPRRN is the only method to successfully detect and remove the long flowing raindrops, even though such a complex raindrop pattern is not included in the training data. This further validates the strong robustness of our DPRRN.
More results on both the synthetic and the real-world datasets can be seen in our supplementary material.

\section{Conclusion}

In this paper, we have proposed the first DP-based raindrop removal network, named DPRRN, consisting of DP raindrop detection and DP fused raindrop removal. In DPRRN, we have utilized DP disparities existing in raindrop regions for robust raindrop detection and fine background detail recovery. We have also constructed both synthetic and real-world DP raindrop removal datasets to train and test our DPRRN. Using those datasets, we have experimentally demonstrated that our DPRRN outperforms existing state-of-the-art methods, showing better generalization ability to real-world raindrops. 
One of our future work is to jointly address rain streaks and raindrops to further boost the real-world applicability.

\vspace{2mm}
\noindent
\textbf{Acknowledgement:} This work was supported by JST SPRING, Grant Number JPMJSP2106.

\bibliography{bmvc_review}
\end{document}